\documentclass[runningheads]{llncs}

\usepackage[T1]{fontenc}
\usepackage{graphicx}
\usepackage{amsmath,amssymb}
\usepackage{multirow}
\usepackage{booktabs}
\usepackage{xcolor}

\begin{document}

\title{RADIANT-PET: Reasoning-Augmented PET/CT Lesion Segmentation with Large Language Models and Reinforcement Learning}
\titlerunning{PET/CT Segmentation with LLM Reasoning}

\author{Jiasheng Wang\inst{1}\and
Tanun Jitwatcharakomol\inst{2}\and
Piyawadee Jongpradubgiat\inst{3}\and
Simeng Zhu\inst{4}}
\authorrunning{J. Wang et al.}
%
\institute{
Division of Hematology, The Ohio State University Comprehensive Cancer Center, Columbus, OH, USA
\and
Department of Radiology, Mahidol University, Bangkok, Thailand
\and
Department of Radiology, Mettapracharak Hospital, Thailand
\and
Department of Radiation Oncology, The Ohio State University Comprehensive Cancer Center, Columbus, OH, USA\\
\email{\{jiasheng.wang,simeng.zhu\}@osumc.edu}
}

\maketitle

\begin{abstract}
Accurate lesion segmentation in PET/CT is critical for oncology, yet remains challenging because physiologic tracer uptake and artifacts can mimic malignant signal. We present RADIANT-PET, a reasoning-augmented framework that couples a high-sensitivity voxel-level segmentation model with lesion-level large language model (LLM) adjudication. Candidate uptake regions are generated with a deliberately permissive segmentation stage, then converted into structured textual descriptions that summarize uptake intensity, morphology, and regional and global anatomical context. An LLM classifies each candidate as true lesion vs. false positive, optionally leveraging the radiology report as additional clinical context. To strengthen lesion-level reasoning, we further optimize a local LLM via reinforcement learning using Group Relative Policy Optimization, rewarding correct lesion classification and anatomically concordant site assignment. Across AutoPET and an OSU test cohort, RADIANT-PET consistently outperforms strong image-only baselines, with the largest improvements observed when radiology reports are provided. Overall, these results demonstrate that LLM-based lesion-level reasoning adds a novel reasoning layer beyond conventional segmentation, suppressing physiologic false positives and aligning voxel-level predictions with clinical interpretation. The project repository is available at: \url{https://github.com/jwang-580/RADIANT-PET}.

\end{abstract}

\section{Introduction}

Positron emission tomography combined with computed tomography (PET/CT) is a cornerstone of oncologic imaging, supporting diagnosis, staging, treatment planning, response assessment, and prognostication~\cite{kostakoglu2019}. Its clinical utility derives from the preferential uptake of radiotracers such as 2-deoxy-2-[\(^{18}\)F]fluoro-D-glucose (FDG) in metabolically active lesions: PET depicts tracer distribution, while CT provides anatomical context. Accurate baseline PET/CT tumor-burden measurement is particularly important in lymphoma because metabolic tumor volume informs prognostic modeling and outcome prediction after therapies such as chimeric antigen receptor (CAR) T-cell therapy. However, physiological uptake, inflammation or post-treatment change, infection, and imaging artifacts can be mistaken for tumor and materially distort these measurements.

Image-only approaches, including nnUNet-based models, have struggled with this task and have not achieved Dice similarity coefficient (DSC) values above 0.8~\cite{kalisch2024,rokuss2024}. In clinical practice, radiologists rely not only on uptake patterns but also on medical reasoning informed by knowledge of physiologic activity, differentiation of inflammation or infection, recognition of technical artifacts, and integration of clinical context. We hypothesize that the performance of purely image-based segmentation models is fundamentally limited by the absence of explicit reasoning capabilities~\cite{lopezpaz2016}. Incorporating the reasoning abilities of large language models (LLMs) offers a principled way to improve discrimination between malignant uptake and physiologic or artifactual activity and thereby improve the reliability of tumor-burden quantification.

\begin{figure}[t]
  \centering
  \includegraphics[width=\textwidth]{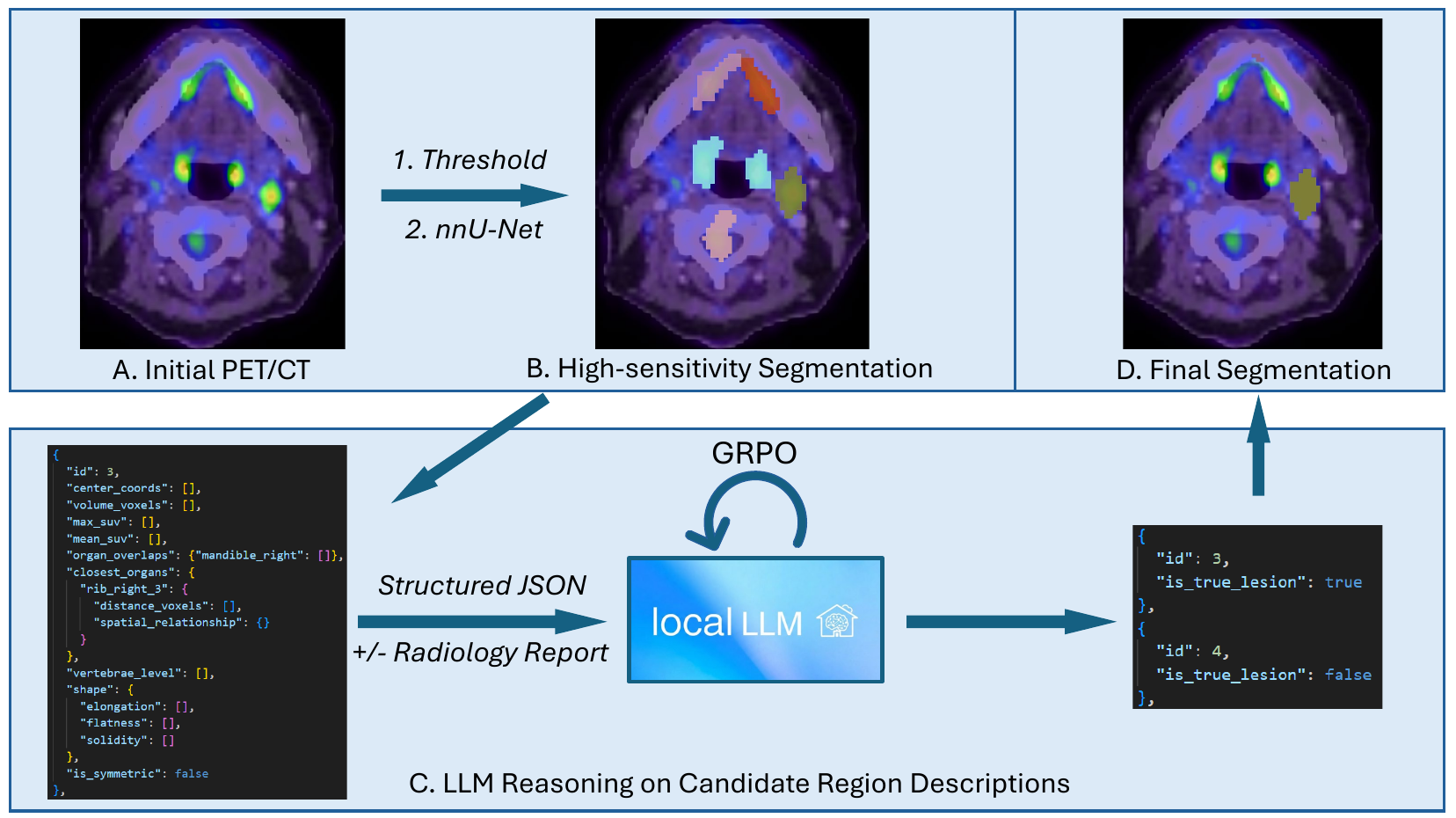}
  \caption{Overview of the RADIANT-PET framework. (A) A high-sensitivity segmenter (SUV thresholding or a modified nnU-Net) proposes a permissive uptake mask. (B) Watershed splitting separates the mask into candidate regions. Each region is then converted into a structured JSON description combining radiomics and organ relationships by incorporating TotalSegmentator masks. (C) A local LLM classifies each candidate as lesion vs. physiological uptake; GRPO reinforcement learning improves this task-specific reasoning. (D) The final segmentation is obtained by removing candidates classified as false positives.}
  \label{fig:overview}
\end{figure}

We propose RADIANT-PET (Reasoning-Augmented Description--Inference Network for PET), a framework that transforms each candidate lesion region into a structured text description including information on tracer uptake characteristics and spatial relationships to surrounding organs. These region-level descriptions are then evaluated by an LLM to classify each uptake as true or false positive. We further enhance the LLM using reinforcement learning with Group Relative Policy Optimization (GRPO), directly optimizing for lesion-level classification accuracy. The framework also natively integrates radiology reports as additional context, enabling joint reasoning over imaging-derived structured descriptions and clinical narrative (Figure~\ref{fig:overview}).

By coupling voxel-level segmentation with lesion-level reasoning, RADIANT-PET achieves strong performance for PET/CT lesion segmentation. Our main contributions are:

\begin{enumerate}
  \item \textbf{Reasoning-augmented PET/CT segmentation framework.} We introduce a hybrid framework that couples conventional segmentation methods for candidate lesion proposal with an LLM-based reasoning stage that adjudicates each candidate.
  \item \textbf{Reinforcement learning for medical reasoning.} We improve LLM lesion adjudication using GRPO-based reinforcement learning with lesion-level supervision.
  \item \textbf{Native integration of radiology reports as physician knowledge.} We directly incorporate free-text radiology reports into the reasoning step, enabling the LLM to natively leverage physician interpretations and clinical context.
\end{enumerate}

\section{Methods}

We adopt a two-stage framework: first, a permissive candidate-generation method (either intensity thresholding or a high-sensitivity nnUNet-based segmenter) proposes all putative uptake regions; second, a reasoning-based large language model adjudicates each candidate using clinical context and anatomical plausibility to retain only true lesions.

\subsection{High-Sensitivity image segmentation model (HS-UNet)}

We train a high-sensitivity segmentation network to generate a deliberately permissive set of candidate lesions, prioritizing recall so that most true lesions are captured at the cost of additional false positives. Our candidate-generation stage follows the model-centric winner of autoPET III (nnUNet-v2), which ranked first on the FDG sub-track~\cite{rokuss2024,kalisch2024}. This model outperformed the earlier autoPET I/II winners based on older nnUNet variants. We consequently use nnUNet-v2 as the state-of-the-art fully automated image-only baseline. It combines large-scale pretraining with organ-supervision--based multitask learning to improve robustness across anatomically diverse uptake patterns.

To prioritize high sensitivity in candidate generation, we modify the training objective. Specifically, the total loss (\(L_{\mathrm{total}}\), Eq.~\ref{eq:total}) is defined as the sum of a Tversky loss (\(L_{\mathrm{Tversky}}\)) and an asymmetric cross-entropy loss (\(L_{\mathrm{ACE}}\)). The Tversky loss~\cite{salehi2017} (Eq.~\ref{eq:tversky}) uses \(\alpha=1\) and \(\beta=0\) to reduce the penalty for false positives, encouraging permissive candidate masks. The ACE term (Eq.~\ref{eq:ace}), motivated by prior work~\cite{wang2019}, further emphasizes false negatives over false positives by setting \(\lambda_{\mathrm{FN}}=8.0\) and \(\lambda_{\mathrm{FP}}=0.125\).

\begin{align}
L_{\mathrm{total}} &= L_{\mathrm{Tversky}} + L_{\mathrm{ACE}}, \label{eq:total}\\
L_{\mathrm{Tversky}} &= 1-\frac{\mathrm{TP}+\epsilon}{\mathrm{TP}+\alpha\cdot\mathrm{FN}+\beta\cdot\mathrm{FP}+\epsilon}, \label{eq:tversky}\\
L_{\mathrm{ACE}} &= \frac{1}{N}\sum_i w_i\left[-\log\frac{e^{z_{i,y_i}}}{\sum_c e^{z_{i,c}}}\right], \label{eq:ace}\\
w_i &= \notag
\begin{cases}
\lambda_{\mathrm{FN}} & \text{if } y_i\ne0 \text{ and } \hat y_i=0,\\
\lambda_{\mathrm{FP}} & \text{if } y_i=0 \text{ and } \hat y_i\ne0,\\
1 & \text{otherwise.}
\end{cases}
\end{align}

During inference, the positive-class decision threshold was lowered from 0.5 to 0.1 to increase sensitivity and generate additional lesion candidates. Aside from the modified loss functions and inference procedure, all other training settings followed the implementation released by Rokuss et al.~\cite{rokuss2024}. We refer to this high-sensitivity nnUNet variant as HS-UNet. The final model is publicly available.

\subsection{Threshold-Based Segmentation and Post-Processing with Watershed Splitting}

We used SUV thresholding as an alternative, model-free approach to identify FDG-avid regions on lymphoma PET. Fixed-SUV thresholding is a clinical-standard approach for lymphoma lesion delineation in international consensus guidance and supports reproducible multicenter measurement~\cite{barrington2014}. Although several rules are used in practice, including SUV \(\geq 2.5\), SUV \(\geq 4.0\), and 40\% of SUVmax, the 2.5 cutoff is validated as a reproducible threshold for lymphoma and is widely used to compute metabolic tumor volume (MTV)~\cite{ilyas2018}. Thresholding is therefore a clinical-standard candidate-generation pipeline. At this stage, the main limitation is not the voxel-level delineation of each FDG-avid region but distinguishing tumor from physiologic and inflammatory uptake at the lesion level. To separate confluent uptake into individual candidate regions, we applied a watershed-based splitting procedure to the resulting masks, for both HS-UNet- and threshold-based segmentations.

\textbf{Stage 1 (distance-based watershed).} For each connected component, we computed the Euclidean distance transform and applied Gaussian smoothing (\(\sigma=2.0\) voxels). Local maxima in the smoothed distance map were used as seed markers with a minimum separation of 8 voxels. We then applied the watershed algorithm to the inverted distance map to generate initial candidate regions.

\textbf{Stage 2 (SUV-valley validation).} To ensure that watershed boundaries corresponded to genuine metabolic separations rather than geometric artifacts, we validated each boundary between adjacent regions. For each neighboring pair \((i,j)\), we computed the relative SUV drop at their shared boundary. We required the relative drop \(\geq 0.60\) for moderate-SUV lesions (peak SUV \(<10\)) and \(\geq 0.75\) for high-SUV lesions (peak SUV \(\geq10\)). If any boundary failed this validation, the split was rejected and the component was retained as a single region.

\subsection{Generation of Candidate Lesion Descriptions}

\begin{figure}[t]
  \centering
  \includegraphics[width=0.92\textwidth]{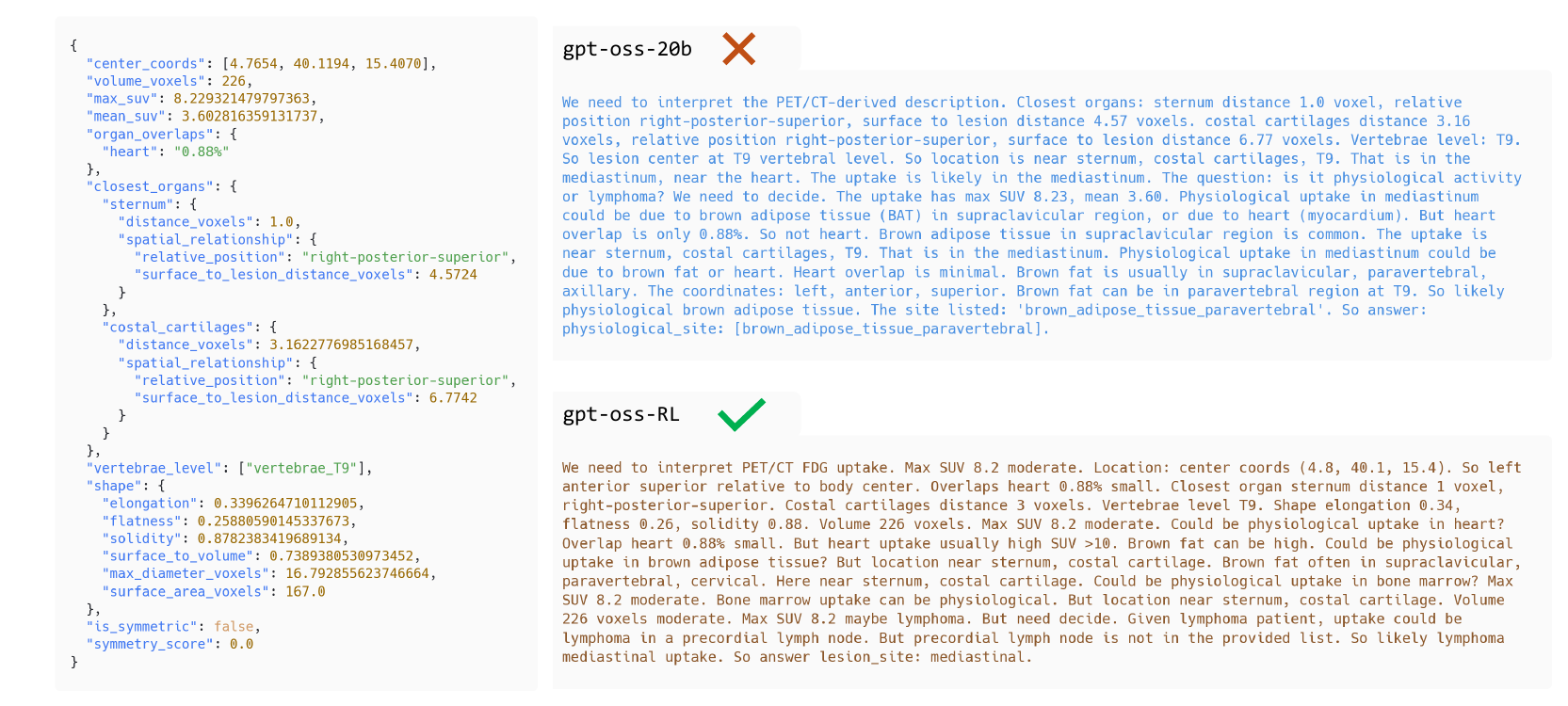}
  \caption{Sample candidate lesion description and model's reasoning trace before and after RL.}
  \label{fig:descriptions}
\end{figure}

Candidate lesion regions were serialized into a structured JSON representation with quantitative imaging features and anatomical information (Figure~\ref{fig:descriptions}). For each uptake, we extracted geometric features (voxel volume), 3D shape descriptors (elongation, flatness, solidity, sphericity, and surface-to-volume ratio), and PET intensity metrics (SUVmax and SUVmean).

Local anatomical context was derived from organ segmentation masks generated by TotalSegmentator~\cite{wasserthal2023}. For each candidate region, we computed (1) the percentage volumetric overlap with each segmented organ and (2) relative-position vectors from the nearest organs' surface points to the uptake's centroid, enabling interpretable spatial descriptions (for example, anterior--superior to the left kidney). Global anatomical context was encoded in two complementary ways. (1) candidate regions were assigned to vertebral levels by matching each uptake's craniocaudal position to the corresponding vertebral body (for example, a mediastinal uptake with centroid height aligned to the T6 vertebral body was labeled as T6 level) and (2) each uptake centroid was referenced to the image-volume center to provide whole-body localization (left/right and superior/inferior relative to body center).

\subsection{GRPO for Uptake Classification}

We applied a reinforcement learning approach to improve classification of candidate lesion regions as either physiological activity or pathological lesions. The method employed Group Relative Policy Optimization (GRPO)~\cite{guo2025} to fine-tune a 20-billion parameter language model (gpt-oss-20b)~\cite{agarwal2025} in BF16 using Low-Rank Adaptation (LoRA). We chose a local LLM over proprietary models to preserve privacy in clinical deployment. gpt-oss-20b was selected because its mixture-of-experts architecture supports fast inference at a relatively small model size and because it showed high specificity at baseline (Table~\ref{tab:classification}). Training used the Unsloth and TRL GRPOTrainer stack~\cite{unsloth2023}. LoRA adapters with rank 8 and \(\alpha=16\) were attached to all attention and MLP projections.

Each candidate lesion region was formatted as structured text that prompted the model to predict both a binary class (\texttt{physiological\_site} vs. \texttt{lesion\_site}) and a specific anatomical site from a vocabulary of 92 predefined locations. GRPO sampled four completions per prompt (group size 4) at temperature 1.0; prompt and completion lengths were each capped at 2,048 tokens. Optimization used AdamW-8bit with a learning rate of \(5\times10^{-5}\), a linear schedule, 10\% warmup, weight decay of \(10^{-3}\), and gradient accumulation of 4. The per-device batch size was 4 for the AutoPET models and 2 for the OSU model.

The composite reward summed three terms. First, binary-label correctness received +1 when the predicted class matched the ground truth and a penalty of \(-1\) if both mutually exclusive labels appeared. Second, anatomical localization received +2 for an exact site match or +1.5 for a match within a curated equivalence group (e.g., bone and bone marrow or related gastrointestinal subregions). Third, a reasoning regularizer applied \(-0.5\) when the reasoning trace did not reference SUVmax, discouraging responses that bypassed imaging features. Reward weights were selected by ablation, with the reported configuration producing the strongest validation F1. We chose GRPO rather than supervised fine-tuning because both the candidate class and anatomical site provide verifiable ground truth suitable for structured rewards; qualitatively, reinforcement learning also produced more clinically grounded reasoning traces (Figure~\ref{fig:descriptions}).

We trained three model variants. Two models were trained on the publicly available AutoPET dataset: one using SUV threshold--based lesion candidates and one using HS-UNet--generated candidates. A third model was trained on the OSU dataset and incorporated radiology reports as additional context, using HS-UNet--generated candidates. All models were trained for 1,200 micro-steps. The AutoPET models used a batch size of 4 (4,800 training samples), whereas the OSU model used a batch size of 2 (2,400 training samples). During inference, predictions were generated deterministically (temperature 0.1), and we evaluated only binary classification (lesion vs. physiological uptake) against the ground truth.

\subsection{Datasets and Annotation}

Four cohorts were used for training and evaluation. \textbf{AutoPET-Train} comprised 120 lymphoma PET/CT cases used for GRPO training; ground-truth lesions were manually assigned to predefined lymph-node stations by a radiologist. \textbf{AutoPET-Test} comprised 20 held-out cases; threshold-based segmentation produced 1,852 candidates (510 true positives and 1,342 false positives), while HS-UNet produced 1,032 candidates (446 true positives and 586 false positives). \textbf{OSU-Train}, collected at The Ohio State University (OSU), comprised 100 lymphoma PET/CT studies with radiology reports used for GRPO training. \textbf{OSU-Test} comprised 20 held-out, de-identified OSU studies; threshold-based segmentation yielded 1,160 candidates (246 true positives and 914 false positives), while HS-UNet yielded 740 candidates (354 true positives and 386 false positives). OSU candidate labels were finalized by a board-certified hematologist and a radiation oncologist. Annotation was performed in 3D Slicer: candidate lesions were manually selected and their boundaries were refined by region expansion or shrinkage to an SUV threshold of 2.5.

\section{Results}

Results are reported at two complementary scales. Table~\ref{tab:classification} evaluates candidate-level classification on the 740 HS-UNet regions from OSU-Test by LLM, whereas Table~\ref{tab:segmentation} evaluates case-level volumetric segmentation and MTV error.

\textbf{High Sensitivity nnUNet.} Modifying training loss of nnUNet produced the intended outcomes (Table~\ref{tab:segmentation}): on OSU-Test, candidate lesions predicted by HS-UNet achieved higher sensitivity than the baseline nnUNet-v2 (0.88 vs. 0.63). Although sensitivity was slightly lower than simple thresholding (0.92), the higher DSC of the HS-UNet (0.58 vs. 0.34) indicates higher-quality candidate lesion predictions.

\textbf{GRPO Improved Local LLM Performance in Candidate Region Classification.} Uptake-level classification performance on OSU-Test candidate regions is summarized in Table~\ref{tab:classification}. GRPO substantially improved the performance of gpt-oss-20b in both report-conditioned and report-free settings.

\begin{table}[t]
\caption{Candidate lesion regions generated by HS-UNet and classified by different LLMs on OSU-Test, with and without radiology report conditioning. gpt-oss-20b-RL denotes gpt-oss-20b fine-tuned with GRPO. ``w/o'' and ``w/'' indicate without and with report conditioning, respectively. For RL models, ``AutoPET''/``OSU'' indicate the RL training dataset.}
\label{tab:classification}
\centering
\begin{tabular}{llcccc}
\toprule
Method & & Sen & Spe & PPV & F1\\
\midrule

\multirow{2}{*}{Gemini-Flash-3.0} 
  & w/o report & 0.87 & 0.28 & 0.52 & 0.65\\
  & w/ report  & 0.93 & 0.49 & 0.63 & 0.75\\
\midrule

\multirow{2}{*}{gpt-5.1} 
  & w/o report & 0.73 & 0.53 & 0.59 & 0.65\\
  & w/ report  & 0.86 & 0.63 & 0.68 & \colorbox{yellow!60}{0.76}\\
\midrule

\multirow{2}{*}{MedGemma} 
  & w/o report & 0.86 & 0.29 & 0.53 & 0.65\\
  & w/ report  & 0.89 & 0.45 & 0.60 & 0.71\\
\midrule

\multirow{2}{*}{gpt-oss-20b} 
  & w/o report & 0.42 & 0.81 & 0.67 & 0.52\\
  & w/ report  & 0.68 & 0.77 & 0.71 & 0.71\\
\midrule

gpt-oss-RL (autoPET) 
  & w/o report & 0.61 & 0.83 & 0.77 & \colorbox{green!30}{0.68}\\
\midrule

gpt-oss-RL (OSU) 
  & w/ report & 0.85 & 0.66 & 0.68 & \colorbox{yellow!60}{0.76}\\

\bottomrule
\end{tabular}
\end{table}

When radiology reports were not provided, the GRPO-tuned model achieved the highest F1 score. When radiology reports were included, GRPO-tuned model achieved performance comparable to the strongest proprietary models.

\textbf{RADIANT-PET Improved Lymphoma Segmentation without Radiology Reports.} Without radiology reports, RADIANT-PET outperformed the nnUNet-v2 baseline on both AutoPET-Test and OSU-Test (Table~\ref{tab:segmentation}). Using threshold-based candidates, the framework with the RL-tuned LLM achieved higher DSC on AutoPET-Test (0.82 vs. 0.78) and OSU-Test (0.77 vs. 0.72). With HS-UNet-generated candidates, the RL-tuned framework again exceeded nnUNet-v2 on OSU-Test (DSC 0.74 vs. 0.72). Overall, RL-tuned lesion-level reasoning matched or exceeded strong image-only segmentation performance even without report conditioning.

\textbf{RADIANT-PET Achieved Best Segmentation Performance with Incorporation of Radiology Reports.} With radiology reports, RADIANT-PET demonstrated the best performance (Table~\ref{tab:segmentation}). Using threshold-based candidates, the framework with the base LLM achieved a DSC of 0.79 and MAPE of 0.57, while the RL-tuned LLM further improved performance to a DSC of 0.82 and MAPE of 0.26. When applied to HS-UNet-generated candidates, the framework incorporating the RL-tuned LLM achieved the best results, with the highest DSC (0.84) and lowest MAPE (0.16). Overall, incorporating radiology reports led to substantial performance gains, with RL providing additional improvements.

\begin{table}[t]
\caption{Segmentation results on AutoPET-Test and OSU-Test. nnUNet-v2 denotes the winning solution of the AutoPET III challenge. +oss and +oss-RL indicate lesion-level LLM adjudication (base gpt-oss-20b vs. GRPO-tuned) applied to filter candidate regions; +Rpt indicates report-conditioned LLM adjudication (OSU-Test only). MAPE (mean absolute percentage error) is computed on metabolic tumor volume (MTV), between predicted and ground-truth total MTV.}
\label{tab:segmentation}
\centering
\resizebox{\textwidth}{!}{%
\begin{tabular}{lcccccccc}
\toprule
& \multicolumn{4}{c}{AutoPET-Test} & \multicolumn{4}{c}{OSU-Test}\\
\cmidrule(lr){2-5}\cmidrule(lr){6-9}
Method & DSC $\uparrow$ & Sen $\uparrow$ & PPV $\uparrow$ & MAPE $\downarrow$ & DSC $\uparrow$ & Sen $\uparrow$ & PPV $\uparrow$ & MAPE $\downarrow$\\
\midrule
Threshold & 0.33 & 0.87 & 0.25 & 13.7 & 0.34 & 0.92 & 0.25 & 16.2\\
nnUNet-v2 & \colorbox{yellow!60}{0.78} & 0.69 & 0.94 & \colorbox{green!30}{0.27} & \colorbox{yellow!60}{0.72} & 0.63 & 0.88 & \colorbox{green!30}{0.83}\\
HS-UNet & 0.59 & 0.92 & 0.46 & 1.90 & 0.58 & 0.88 & 0.45 & 2.05\\
\midrule
Threshold+oss & 0.69 & 0.72 & 0.80 & 2.44 & 0.50 & 0.42 & 0.77 & 0.84\\
Threshold+oss-RL & \colorbox{yellow!60}{0.82} & 0.85 & 0.85 & \colorbox{green!30}{0.20} & \colorbox{yellow!60}{0.77} & 0.73 & 0.85 & \colorbox{green!30}{0.27}\\
HS-UNet+oss & 0.45 & 0.41 & 0.54 & 0.72 & 0.47 & 0.38 & 0.77 & 0.78\\
HS-UNet+oss-RL & 0.75 & 0.84 & 0.70 & 0.64 & 0.74 & 0.71 & 0.83 & 0.51\\
\midrule
Threshold+Rpt+oss & \multicolumn{4}{c}{--} & 0.79 & 0.85 & 0.81 & 0.57\\
Threshold+Rpt+oss-RL & \multicolumn{4}{c}{--} & 0.82 & 0.86 & 0.81 & 0.26\\
HS-UNet+Rpt+oss & \multicolumn{4}{c}{--} & 0.79 & 0.77 & 0.85 & 0.22\\
HS-UNet+Rpt+oss-RL & \multicolumn{4}{c}{--} & \colorbox{yellow!60}{0.84} & 0.85 & 0.84 & \colorbox{green!30}{0.16}\\
\bottomrule
\end{tabular}}
\end{table}

\section{Discussion and Conclusion}

In this work, we introduced RADIANT-PET, a reasoning-augmented framework that combines high-sensitivity voxel-level segmentation with lesion-level adjudication using LLMs. By decoupling candidate generation from final decision-making, the approach addresses a key limitation of image-only PET/CT segmentation: the lack of explicit reasoning about physiological uptake, anatomy, and clinical context. Lesion-level reasoning effectively suppresses false positives while preserving sensitivity, leading to improved segmentation accuracy and metabolic tumor volume estimation.

Several findings merit discussion. First, reinforcement learning with GRPO improved the performance of a local LLM, narrowing the gap with proprietary models and enabling deployment in privacy-sensitive clinical environments. Second, incorporating radiology reports provided a strong complementary signal, allowing the model to align image-derived findings with physician interpretation and yielding the best overall performance. Importantly, even without report conditioning, RL-tuned lesion-level reasoning matched or exceeded strong image-only baselines, suggesting that explicit reasoning adds value beyond additional clinical context. These gains directly address clinically consequential false positives from physiologic uptake, inflammation or post-treatment change, infection, and artifacts that can distort MTV and downstream prognostic models.

During development, we also evaluated the medical vision-language model MedGemma, which showed weak lesion-level classification, likely because current open-source medical VLM pretraining contains limited PET/CT data. Effective VLM fine-tuning may require substantially larger multimodal datasets. A per-lesion patch CNN is another useful future baseline, but unlike the present framework it cannot natively condition on the radiology report, a central input and a major source of the observed performance gains.

This study is limited by its focus on lymphoma FDG PET/CT and reliance on upstream organ segmentation and handcrafted features. Future work will extend the framework to additional tracers and disease types, investigate multimodal fine-tuning, and compare against dedicated per-lesion image classifiers. 

In summary, RADIANT-PET demonstrates that LLM-based lesion-level reasoning can serve as an effective post-segmentation adjudication layer, providing an interpretable approach for clinically aligned PET/CT lesion segmentation.

\begin{credits}
\subsubsection{\discintname}
The authors have no competing interests to declare that are
relevant to the content of this article.
\end{credits}

\end{document}